\documentclass{article}
\usepackage[preprint]{log_2024}			
\usepackage{subfig}
\usepackage{amsthm}
\newtheorem{definition}{Definition}[section]
\newtheorem{theorem}{Theorem}[section]
\usepackage{booktabs}						
\usepackage{multirow}						
\usepackage{amsfonts}						

\usepackage[numbers,compress,sort]{natbib}	

\usepackage{bm}

\renewcommand{\top}{{\mkern-1.5mu\mathsf{T}}}

\newcommand{\be}{\begin{equation}}
\newcommand{\ee}{\end{equation}}
\newcommand{\beq}{\begin{equation}}
\newcommand{\eeq}{\end{equation}}
\newcommand{\beqy}{\begin{eqnarray}}
\newcommand{\eeqy}{\end{eqnarray}}
\newcommand{\beqynn}{\begin{eqnarray*}}
\newcommand{\eeqynn}{\end{eqnarray*}}

\newcommand{\ba}{\begin{array}}
\newcommand{\ea}{\end{array}}
\newcommand{\bmx}{\begin{bmatrix}}
\newcommand{\emx}{\end{bmatrix}}
\newcommand{\bsmx}{\left[\begin{smallmatrix}}
\newcommand{\esmx}{\end{smallmatrix}\right]}
\newcommand{\bmxc}[1]{\left[\begin{array}{@{}#1@{}}}
\newcommand{\emxc}{\end{array}\right]}

\newcommand{\bt}[1]{\begin{tabular}{#1}}
\newcommand{\et}{\end{tabular}}

\newcommand{\bc}{\begin{center}}
\newcommand{\ec}{\end{center}}
\newcommand{\ben}{\begin{enumerate}}
\newcommand{\een}{\end{enumerate}}
\newcommand{\bi}{\begin{itemize}}
\newcommand{\ei}{\end{itemize}}

\usepackage{scalerel}

\newcommand{\diag}{\mathrm{diag}}

\newcommand{\calN}{\mathcal{N}}

\newcommand{\Rbb}{{\mathbb{R}}}

\newcommand{\A}{\mathbf{A}}
\newcommand{\B}{\mathbf{B}}
\newcommand{\C}{\mathbf{C}}
\newcommand{\D}{\mathbf{D}}

\newcommand{\I}{\mathbf{I}}

\renewcommand{\P}{\mathbf{P}}

\newcommand{\U}{\mathbf{U}}

\newcommand{\W}{\mathbf{W}}

\newcommand{\Y}{{\mathbf{Y}}}
\newcommand{\Z}{\mathbf{Z}}

\newcommand{\f}{\mathbf{f}}

\newcommand{\h}{\mathbf{h}}

\newcommand{\x}{{\mathbf{x}}}
\newcommand{\y}{{\mathbf{y}}}
\newcommand{\z}{{\mathbf{z}}}

\newcommand{\phib}{\bm{\phi}}

\newcommand{\bLambda}{\boldsymbol{\Lambda}}

\newcommand{\0}{{\boldsymbol{0}}}
\newcommand{\1}{{\boldsymbol{1}}}

\newcommand{\rw}{{\mathrm{rw}}}
\newcommand{\sym}{{\mathrm{sym}}}

\newcommand{\red}{\textcolor{red}}
\newcommand{\blue}{\textcolor{blue}}

\newcommand\ie{\textit{i.e.,}}

\title[Flexible Diffusion Scopes with Parameterized Laplacian for Heterophilic Graph Learning]{Flexible Diffusion Scopes with Parameterized Laplacian for Heterophilic Graph Learning}

\author{
Qincheng Lu$^{1}$,  Jiaqi Zhu$^{1}$, Sitao Luan$^{1, 2,*}$,  Xiao-Wen Chang$^{1,*}$ \\
\{qincheng.lu@mail, sitao.luan@mail, jiaqi.zhu@mail, chang@cs\}.mcgill.ca\\
$^1$McGill University; $^2$ Mila - Quebec Artificial Intelligence Institute;  $^*$Corresponding Author
}

\begin{document}

\maketitle

\begin{abstract}

The ability of Graph Neural Networks (GNNs) to capture long-range and global topology information is limited by the scope of conventional graph Laplacian, leading to unsatisfactory performance on some datasets, particularly on heterophilic graphs. To address this limitation, we propose a new class of parameterized Laplacian matrices, which provably offers more flexibility in controlling the diffusion distance between nodes than the conventional graph Laplacian, allowing long-range information to be adaptively captured through diffusion on graph. Specifically, we first prove that the diffusion distance and spectral distance on graph have an order-preserving relationship. With this result, we demonstrate that the parameterized Laplacian can accelerate the diffusion of long-range information, and the parameters in the Laplacian enable flexibility of the diffusion scopes. Based on the theoretical results, we propose topology-guided rewiring mechanism to capture helpful long-range neighborhood information for heterophilic graphs. With this mechanism and the new Laplacian, we propose two GNNs with flexible diffusion scopes: namely the Parameterized Diffusion based Graph Convolutional Networks (PD-GCN) and Graph Attention Networks (PD-GAT). Synthetic experiments reveal the high correlations between the parameters of the new Laplacian and the performance of parameterized GNNs under various graph homophily levels, which verifies that our new proposed GNNs indeed have the ability to adjust the parameters to adaptively capture the global information for different levels of heterophilic graphs. They also outperform the state-of-the-art (SOTA) models on 6 out of 7 real-world benchmark datasets, which further confirms their superiority.

\end{abstract}
\vspace{-0.2cm}
\section{Introduction}

Combining graph signal processing and Convolutional Neural Networks (CNNs)~\cite{lecun1998gradient}, Graph Neural Networks (GNNs) has achieved remarkable success on machine learning tasks with non-Euclidean data~\cite{scarselli2008graph,kipf2016semi,
hamilton2017inductive,velivckovic2017graph,xu2018powerful,luan2019break,beaini2021directional,zhao2021consciousness,hua2023mudiff,lu2024gcepnet}. In GNNs, high-order neighbors are recursively incorporated through diffusion across multiple stacked layers. In each layer, unlike CNNs where neighbors are weighted differently, the convolutional kernel in the vanilla GNNs~\cite{kipf2016semi} and many other popular variants~\cite{hamilton2017inductive, xu2018powerful} assign the same weight in a neighbourhood, or use weights determined by commonly used normalization of adjacency or Laplacian matrices.


Recent studies show that the optimal choice of a normalized Laplacian is data-dependent~\cite{dasoulas2021learning, dall2020optimal}. Aggregation with learnable weights has been proposed to enrich GNNs expressiveness~\cite{eliasof2023improving}. However, limited knowledge about long-range neighbors and the global graph structure prevents conventional local aggregation from achieving optimal performance~\cite{gasteiger2019diffusion}. For example,
while GAT~\cite{velivckovic2017graph} allows for learnable weights, it generates node representations solely based on the representations of its direct neighbours.  This works well on the homophilic graphs~\cite{luan2022we, luan2024graph}, where a node and its neighbours are likely to have the same label.  However, as the limited and fixed diffusion scope, GAT suffers from significant performance loss in node classification tasks involving heterophilic graphs~\cite{zhu2020beyond,luan2022complete}, where nodes from different classes tend to be connected.  

Non-local neighborhood information is found to be helpful to deal with heterophily problem for GNNs~\cite{abu2019mixhop,liu2020non,zhu2020beyond,he2021bernnet,luan2024heterophilic}, but it has not been fully explored considering the design of graph Laplacian and its diffusion scope.
We will address this issue in this paper. The main contributions of our work are as follows: 
\textbf{(1)} We propose a new class of parameterized normalized graph Laplacian matrices, which offer better control over the diffusion process, and include several widely used normalized Laplacians as special cases. 
Through the parameters in new Laplacian, we can adjust the diffusion and spectral distances between nodes by altering spectral properties of the graph, and thus enable flexible message passing to capture local and global information adaptively.
\textbf{(2)} We establish a theorem to prove that spectral distance can be used as a surrogate function of diffusion distance, which significantly reduces the computational cost for comparing relative distance between nodes. The theoretical results substantially extends the scope of~\cite{beaini2021directional} and overcomes some of its shortcomings.
\textbf{(3)} Based on the new graph Laplacian and the theorem, we propose two architectures, the Parameterized Diffusion based Graph Convolutional Networks (PD-GCN) and Graph Attention Networks (PD-GAT), which enable flexible diffusion.
The empirical results demonstrate the effectiveness and superiority of the proposed models compared with SOTA GNNs for node classification tasks on graph across various homophily levels, especially on heterophilic graphs. The proposed strategy is characterized by its flexibility and seamless integration with other types of graph aggregation.

This paper is organized as follows:  notation and background knowledge are introduced in Section \ref{sec:preliminaries}; in Section \ref{sec:DAM}, we propose the new class of Laplacian, show its properties, prove the theoretical results on spectral and diffusion distances and present the proposed architectures; 
in Section \ref{sec:experiments}, we show the experimental results and comparisons on synthetic and real-world graphs, and conduct ablation study.


\vspace{-0.3cm}
\section{Preliminaries}
\label{sec:preliminaries}

In this section, we introduce notation and background knowledge.
We use {\bf bold} font for vectors and matrices.
For a matrix $\B=(b_{ij})$, $|\B|=(|b_{ij}|)$
and its $i^{th}$ row is denoted as $\B_{i,:}$ or $\mathbf{b}_i^\top$.
We use $\B || \C$ and $[\B, \C]$ to denote column and row concatenation of matrices $\B$ and $\C$, respectively.
We use $\mathcal{G}=(\mathcal{V}, \mathcal{E})$ to denote an undirected connected graph with the vertex set $\mathcal{V}$ (with $|\mathcal{V}| = N$) and the edge set $\mathcal{E}$.
We have a node feature matrix $\mathbf{X} \in \Rbb^{N \times d_{0}}$ whose $i^{th}$ row is the transpose of the feature vector $\mathbf{x}_i \in \Rbb^{d_{0}}$ of node $v_i$.
The learned node representation matrix 
at the $l^{th}$ layer of GNNs is denoted 
by $\mathbf{H}^{(l)} \in \Rbb^{N \times d_{l}}$.
For a node $v_i \in \mathcal{V}$, 
$\mathcal{N}(v_i) \subseteq \mathcal{V}$ denotes the set of neighbouring nodes of $v_i$.
The adjacency matrix of $\mathcal{G}$ is denoted by $\mathbf{A}=(a_{ij}) \in \Rbb^{N \times N}$ with $a_{ij} = 1$ if $e_{ij} \in \mathcal{E}$ and $a_{ij} = 0$ otherwise.
The degree matrix of $\mathcal{G}$ is 
$\mathbf{D}=\diag(d_{ii})\in \Rbb^{N \times N}$
with $d_{ii} = \sum_{j} a_{ij}$.
Three Laplacian matrices, namely the combinatorial Laplacian, random-walk normalized Laplacian and symmetric normalized Laplacian are respectively defined as:
\be 
\label{def:lap}
\mathbf{L} = \mathbf{D} - \mathbf{A}, \ \ \ \mathbf{L}_{\mathrm{rw}} = \mathbf{D}^{-1} \mathbf{L},  \ \ \ \mathbf{L}_\mathrm{sym} = \mathbf{D}^{-\frac{1}{2}} \mathbf{L} \mathbf{D}^{-\frac{1}{2}}.
\ee
We use $\boldsymbol{\phi}^{(k)}$ to denote the $k^{th}$ eigenvector of a Laplacian corresponding to the $k^{th}$ smallest positive eigenvalue $\lambda^{(k)}$.

\subsection{Local Graph Aggregation}
Most modern GNNs adopt the message-passing framework~\cite{hamilton2020graph}, 
in which the representation $\mathbf{h}_u$ of node $u$ is generated by iterative local aggregation of its neighbors and its own representation from the previous layer~\cite{hamilton2020graph}. 
The local graph aggregation at the $l^{th}$ layer is expressed as:
\be
\mathbf{H}^{(l)} = \sigma ( \mathbf{S} \mathbf{H}^{(l-1)} \mathbf{W}^{(l)}),
\ee
where $\mathbf{W}^{(l)} \in \Rbb^{d_{l} \times d_{l-1}}$ is a learnable parameter and $\sigma (\cdot)$ is a non-linear activation function.
Here $\mathbf{S}$ is the aggregation operator that provides weights for neighboring messages.
Common choices for a fixed $\mathbf{S}$ include adjacency matrices corresponding to $\mathbf{L}$, $\mathbf{L}_{\mathrm{rw}}$, and $\mathbf{L}_\mathrm{sym}$. 
Alternatively, learnable weights can be assigned.
One approach to assigning learnable weights is graph attention~\cite{lee2019attention},
where, $r_{ij}^{(l)}$, the unnormalized attention score that node $v_i$ gives to node $v_j$ at the $l$-th layer is defined as
\be
\label{eq:leaky_attention}
r_{ij}^{(l)} = \text{LeakyReLU}\big((\mathbf{a}^{(l)})^{\top} [\mathbf{W}^{(l)} \mathbf{h}_i^{(l-1)} || \mathbf{W}^{(l)} \mathbf{h}_j^{(l-1)}]\big),
\ee
where $\mathbf{a}^{(l)} \in \Rbb^{2 d_{l}}$ is trainable.
The aggregation at the $l$-th layer is weighted by $\mathbf{S}_{ij}^{(l)} = \text{softmax}_j(r_{ij}^{(l)})$.



\subsection{Spectral-based Nodes Relative Distances}
The relative position of nodes on the graph could be told by eigenvectors of Laplacian matrices~\cite{kreuzer2021rethinking}. As the graph data is non-Euclidean, various measurements are proposed to describe the distance between nodes~\cite{nadler2005diffusion, Qiu2007, Belkin2003}.
Among them,
the diffusion distance is often used to model how node $v_i$ influence node $v_j$  by considering random walks along edges~\cite{beaini2021directional}.
At the initial step,
the random walk starts at a node, 
and moves to one of its neighbours at the next step.
The diffusion distance between $v_i$ and $v_j$ is proportional to the probability that the random walk starting at node $v_i$ meets the random walk starting at node $v_j$ at step $t$.
Based on $\mathbf{L}_\rw$,
the \textbf{diffusion distance} is calculated as (see~\cite{Coifman2006Diffusion}):
\be
\label{def:diffusion-dist}
d_t(v_i, v_j; \mathbf{L}_\rw) = \Big( \sum_{k=1}^{n-1} e^{-2t \lambda^{(k)}}(\phi_i^{(k)} - \phi_j^{(k)})^2 \Big)^\frac{1}{2}.
\ee
Another measure of the distance between $v_i$ and $v_j$ 
is the \textbf{spectral distance} defined as (see~\cite{Belkin2003}):
\be
\label{def:spectral-dist}
d_{s}(v_i, v_j; \mathbf{L}_\rw) = | \phi_i^{(1)}- \phi_j^{(1)}|.
\ee
Note that $\phib^{(1)}$ gives positional information of nodes~\cite{dwivedi2020generalization, kreuzer2021rethinking, muller2023attending}.
A more general definition of $d_s$ involves the $k$ eigenvectors
corresponding to the $k$ smallest eigenvalues,
but here we take $k=1$ for simplicity and informativeness.  

\subsection{Related Works}
\label{subsec:dgn}
\paragraph{Design of Graph Laplacian and Adjacency Matrices} Efforts have been made to overcome the drawbacks of message passing with the conventional graph Laplacian or adjacency matrices.
This includes works that incorporate novel attention mechanisms into GNNs~\cite{Brody2021How, yun2019graph, hu2020heterogeneous, wang2020direct,luan2022revisiting},
as well as transformer-based GNNs that consider fully connected graphs~\cite{kreuzer2021rethinking, muller2023attending}. The parameterized graph shift operator (PGSO)~\cite{dasoulas2021learning} generalizes conventional adjacency matrices, while $\omega$GNN~\cite{eliasof2023improving} offers trainable weighting factors for aggregation. The Directional Graph Network (DGN)~\cite{beaini2021directional} proposes aggregation matrices, namely the directional average matrix $\mathbf{B}_{\text{av}}$ and the directional derivative matrix $\mathbf{B}_{\text{dx}}$, which contain weights that represent the topological importance based on eigenvectors of the conventional Laplacian. Based on a given eigenvector $\boldsymbol{\phi}$, they are defined as:
\be
\label{eq:directedagg}
\mathbf{B}_{\mathrm{av}}(\boldsymbol{\phi}) \!=\!  |\nabla \boldsymbol{\phi}|, \; 
\mathbf{B}_{\mathrm{dx}}(\boldsymbol{\phi}) \!=\! \nabla \boldsymbol{\phi} - \text{diag}(\nabla \boldsymbol{\phi} \mathbf{1}), \:
\nabla\phi_{ij} = \phi_j - \phi_i \: \: \text{if} \: \: e_{ij} \in \mathcal{E} \: \: \text{else} \:  \: 0.
\ee
The proposed flexible diffusion with the new class of parameterized Laplacian matrices in our work differs from these works as it offers a theoretical way to control the spectral properties of graphs and allows adaptation to different homophily levels.
Our proposed method enhances conventional message passing,
and is independent of the specific local aggregation framework or attention mechanism it is combined with, offering the potential to improve various GNNs, including graph transformers.

\paragraph{Heterophily and Long-range Topology Information} On heterophily graphs, nodes tend to connect others that have different labels. 
Regarding to heterophilic graph learning, recent studies have found that the global receptive field is more desired and necessary than local information~\cite{xing2024less, luan2024heterophilic}, as the information from the same class are more likely to come from distant nodes. However, GNNs will suffer from information loss with an increased number of diffusion steps, a phenomenon known as over-squashing~\cite{alon2020bottleneck}.
Consequently, the difficulty in obtaining long-range neighborhood information without compromising its quality explains the unsatisfactory performance of GNNs on heterophilic graphs.
Meanwhile, transformer-based GNNs
still fall short on existing heterophilic benchmarks~\cite{muller2023attending},
suggesting that naively creating edges between distant nodes cannot handle this issue.
Various rewiring methods also have been proposed~\cite{rong2019dropedge, papp2021dropgnn, abu2019mixhop, arnaiz2022diffwire, alon2020bottleneck}. Our work addresses the learning challenge on heterophilic graphs by accelerating diffusion from long-range neighbors, and is characterized by both the flexibility of the parameterized diffusion and the topology-guided rewiring strategy.

\vspace{-0.3cm}
\section{Parameterized Normalized Laplacian and Flexible Diffusion}
\label{sec:DAM}
We first propose a new parameterized normalized Laplacian matrix, 
which defines a new class of graph Laplacians with its directional information that adapts to the graph's homophily levels,
as described in Section~\ref{sec:para-lap}. 
Then we discuss how this new Laplacian enables local aggregation to become aware of the global graph structure.
Specifically,
we propose two variants that incorporate the new parameterized normalized Laplacian into GNNs in Section~\ref{sec:pd}:
the Parameterized Diffusion based Graph Convolutional Networks (PD-GCN) 
and Graph Attention Networks (PD-GAT). 
Based on the new Laplacian, 
we propose the topology-guided graph rewiring strategy,
and its ability to capture long-range neighborhood information in heterophily graphs is theoretically justified in Section~\ref{sec:rewire}.


\subsection{Parameterized Normalized Laplacian}
\label{sec:para-lap}
In order to gain a more refined control over the diffusion on a graph, we define a new class of Laplacian matrices, denoted by $\mathbf{L}^{(\alpha,\gamma)}$, where the two parameters $\gamma$ and $\alpha$ determine its spectral properties. Then, we propose an efficient method to compare the relative diffusion distances.

\begin{definition}
\label{def:aug-lap}
A parameterized normalized Laplacian matrix is defined as
\be \label{eq:plaplacian}
\mathbf{L}^{(\alpha,\gamma)} 
= \gamma [\gamma \D + (1-\gamma )\I]^{-\alpha} \mathbf{L}
[\gamma \D + (1-\gamma )\I]^{\alpha-1}
\ee
and the corresponding parameterized normalized adjacent matrix  
is defined as
\be \label{def:adj}
\P^{(\alpha,\gamma)} =  \I-\mathbf{L}^{(\alpha,\gamma)},
\ee
where the parameters $\gamma \in (0, 1]$ and $\alpha \in [0, 1]$. 
\end{definition}

The normalized Laplacian matrices defined in Eq.~\eqref{def:lap} are special cases of this new class of Laplacian matrices: when $\alpha=1$ and $\gamma=1$, $\mathbf{L}^{(\alpha,\gamma)}=\mathbf{L}_\rw$; when $\alpha=\frac{1}{2}$ and $\gamma=1$, $\mathbf{L}^{(\alpha,\gamma)}=\mathbf{L}_\sym$.  
Although we cannot choose $\alpha$ and $\gamma$ such that 
$\mathbf{L}^{(\alpha,\gamma)}$ becomes $\mathbf{L}$, 
we have the following result:
\be 
\begin{aligned}
\label{proof:inf}
\lim_{\gamma \to 0} \frac{1}{\gamma}\mathbf{L}^{(\alpha, \gamma)} = \lim_{\gamma \to 0} (\gamma \D + (1-\gamma )\I)^{-\alpha} \mathbf{L} \times \lim_{\gamma \to 0} (\gamma \D + (1-\gamma )\I)^{\alpha-1} = \mathbf{L}, 
\end{aligned}
\ee
which implies that when $\gamma$ is small enough, 
an eigenvector of $\mathbf{L}^{(\alpha,\gamma)}$ is a good approximation to
an eigenvector of $\mathbf{L}$.
The following theorem justifies the definition of $\P^{(1,\gamma)}$ as a random walk matrix,
which generalizes the classic random walk matrix $\D^{-1}\A$.
\begin{theorem}
\label{thm:parameterized-matrix}
The $\P^{(\alpha, \gamma)}$ defined in \eqref{def:adj} is non-negative (\ie{} all of its elements are non-negative), and when $\alpha=1$, $\P^{(\alpha, \gamma)}\1= \1$. 
See the proof in Appendix \ref{appendix:proof_parameterized_matrix}.
\end{theorem}
The properties of eigenvalues of $\mathbf{L}^{(\alpha,\gamma)}$ 
are given in the following Theorem.

\begin{theorem}
\label{thm:eigenplaplaican}
    Suppose the graph $\mathcal{G}$ is connected.
    Then the symmetric $\mathbf{L}^{(1/2,\gamma)}\in \Rbb^{n\times n}$ has the eigendecomposition:
    \be \label{eq:gled}
    \mathbf{L}^{(1/2,\gamma)} = \U \bLambda^{(\gamma)} \U^\top,
    \ee
    where $\U\in \Rbb^{N\times N}$ is orthogonal and $\bLambda^{(\gamma)}=\diag(\lambda^{(i)}(\gamma))$,
    \be \label{eq:glevrange}
    0 =\lambda^{(0)}(\gamma) < \lambda^{(1)}(\gamma)
    \leq \cdots \leq \lambda^{(N-1)}(\gamma) \leq 2.
    \ee
Each $\lambda^{(i)}(\gamma)$ is strictly increasing with respect to $\gamma$ for $i=1:N-1$.
Furthermore, $\mathbf{L}^{(\alpha,\gamma)}$ has the eigendecomposition
\be \label{eq:gled-alpha}
\mathbf{L}^{(\alpha,\gamma)} 
= \left([\gamma\D+(1-\gamma)\I]^{\frac{1}{2}-\alpha} \U \right) \bLambda^{(\gamma)} 
\left([\gamma\D+(1-\gamma)\I]^{\frac{1}{2}-\alpha}\U\right)^{-1} ,
\ee
\ie{} $\mathbf{L}^{(\alpha,\gamma)}$ share the same eigenvalues as $\mathbf{L}^{(1/2,\gamma)}$ and the columns of $[\gamma\D+(1-\gamma)\I]^{\frac{1}{2}-\alpha}\U$ are the corresponding eigenvectors. See proof in Appendix~\ref{appendix:proof_eigenplaplaican}.
\end{theorem}

The following theorem shows the monotonicity of the diffusion distance with respect to the spectral distance under certain conditions. Here both distances are defined in terms of eigenvectors of 
$\mathbf{L}^{(1,\gamma)}$,
cf.\ \eqref{def:diffusion-dist} and \eqref{def:spectral-dist},
which involve the eigenvectors of $\mathbf{L}_\rw$.

\begin{theorem}
\label{thm:gradient-v2}
Let $v_i$, $v_j$ and $v_m$ be nodes of the graph $\mathcal{G}$ such that 
$d_{s}(v_m, v_j; \mathbf{L}^{(1,\gamma)}) < d_{s}(v_i, v_j; \mathbf{L}^{(1,\gamma)})$.
Then there is a constant $C$ such that for $t\geq C$,  
\be
    d_t(v_m, v_j; \mathbf{L}^{(1,\gamma)}) < d_t(v_i, v_j; \mathbf{L}^{(1,\gamma)}).
\ee
Furthermore, the ratio $d_t(v_m, v_j; \mathbf{L}^{(1,\gamma)})/d_t(v_i, v_j; \mathbf{L}^{(1,\gamma)})$
is proportional to $e^{-\lambda^{(1)}(\gamma)}$. 
\end{theorem}

Theorem \ref{thm:gradient-v2}\footnote{Theorem \ref{thm:gradient-v2} substantially extends the scope of~\cite{beaini2021directional}, which deals with the diffusion distance defined by the random walk Laplacian $\mathbf{L}_{\rw}$. Furthermore, Theorem \ref{thm:gradient-v2} overcomes some shortcomings of~\cite{beaini2021directional}, see details in Appendix~\ref{appendix:gradient-v2}.} shows that the spectral distance can be used as a good indicator of the diffusion distance, \ie{} if node $i$ has a larger spectral distance to node $j$ than node $m$, then node $i$ would also have a larger diffusion distance to node $j$ after enough time steps. According to \eqref{def:diffusion-dist}, we needs to find all eigenvalues and eigenvectors of the Laplacian to calculate the diffusion distance, which is computationally expensive. But as Theorem \ref{thm:gradient-v2} shows, we can easily compute the spectral distance and use it as a surrogate function of diffusion distance, which is much more efficient.
Since a smaller diffusion distance implies greater influence between two nodes in GNNs,
DGN~\cite{beaini2021directional} defines the directional aggregation operator that embeds the diffusion distance. 
However, 
DGN~\cite{beaini2021directional} requires $k$ eigenvectors of $\mathbf{L}_{\rw}$,
where a larger $k$ is preferred for better capturing of the actual diffusion process.
While with Theorem \ref{thm:gradient-v2},
we can encode such global structural information directly into the local aggregation by utilizing the spectral distance,
which only involves the first non-trivial eigenvectors.




Furthermore,
$\mathbf{L}^{(\alpha,\gamma)}$ allows for nuanced control over message passing.
For graphs with different homophily levels, the node would prefer neighborhood information from different diffusion distances. Specifically, nodes with small diffusion distance would be favored on homophilic graphs and nodes with large diffusion distance would be helpful on heterophilic graphs to reduce the effect of noises~\cite{topping2021understanding}.
Adjusting $\alpha$ and $\gamma$ can change the eigenvalues and eigenvectors of  $\mathbf{L}^{(\alpha,\gamma)}$, 
thereby altering diffusion distance. 
For instance, we can employ a smaller $\gamma$ on heterophilic graphs to help the propagation of long-range neighborhood information and employ a larger $\gamma$ on homophilic graphs to let GNNs to focus on useful local information. 
Therefore, 
$\alpha$ and $\gamma$ can be tuned as hyperparameters
to enhance GNNs expressiveness according to the homophily levels.

\subsection{Parameterized Diffusion Augmented GNNs}
\label{sec:pd}
We present the proposed methods, which aim to adapt the diffusion process to the graph homophily levels. Two GNNs, the Parameterized Diffusion based GCN (PD-GCN) and GAT (PD-GAT), achieve such flexible diffusion through the new parameterized normalized Laplacian $\mathbf{L}^{(\alpha, \gamma)}$. These paradigms overcome the limitations of message passing on the graph structures encoded by the conventional Laplacian, and can accelerate the diffusion of long-range information in heterophilic graphs.

\textbf{PD-GCN}: The one-hop neighborhood aggregation in GCN~\cite{kipf2016semi} is a first-order approximation of the graph convolution using a polynomial filter on the graph Laplacian~\cite{defferrard2016convolutional}.
Accordingly, we propose Parameterized Diffusion based Graph Convolutional Networks (PD-GCN), 
where the aggregation utilizes the parameterized normalized adjacent matrix $\P^{(\alpha,\gamma)} = \I - \mathbf{L}^{(\alpha,\gamma)}$ instead.
In PD-GCN, the $l^{th}$ layer computes the updated representation as:
\be
\mathbf{H}^{(l)} = \mathrm{\sigma} (\P^{(\alpha,\gamma)} \mathbf{H}^{(l-1)} \mathbf{W}^{(l)}),
\ee
here $\P^{(\alpha,\gamma)}$ serves as weights for aggregating neighboring nodes. While the vanilla GCN sets the weights solely based on paired nodes degrees, PD-GCN is able to encode global information into the local aggregation with its flexible diffusion scopes, surpassing the expressiveness of GCN for the following reasons: The first non-trivial eigenvector of the graph Laplacian tells the diffusion distance between nodes, as shown in Theorem \ref{thm:gradient-v2}, and thus indicates the dominant direction of diffusion. However, whether a graph is homophilic or heterophilic is independent of its spectral properties. As a result, the diffusion governed by the conventional Laplacian may not be optimal. In $\mathbf{L}^{(\alpha,\gamma)}$, with suitable values of $\alpha$ and $\gamma$ that account for the graph's homophily levels, PD-GCN can adjust its diffusion scopes accordingly to perform more effective message propagation.

\textbf{PD-GAT}:
The learnable aggregation weights in GAT are merely based on local node features, and does not adapt to the specific characteristics of the global topology. 
PD-GAT addresses this shortcoming by equipping the attention mechanism with parameterized edge features.
Denote the first non-trivial eigenvector of $\mathbf{L}^{(\alpha, \gamma)}$ as $\phib^{(1)}(\alpha, \gamma)$,
the feature of $e_{ij} \in \mathcal{E}$ is defined as follows:
\be
\label{eq:edgefeature_dgn}
\f^{(i,j)} (\alpha, \gamma) = \big[ \mathbf{B}_{\mathrm{av}}(\phib^{(1)}(\alpha, \gamma))_{ij}, \mathbf{B}_{\mathrm{dx}}(\phib^{(1)}(\alpha, \gamma))_{ij} \big]^\top,
\ee
where 
$\mathbf{B}_{\mathrm{av}}(\phib^{(1)}(\alpha, \gamma))$ and $\mathbf{B}_{\mathrm{dx}}(\phib^{(1)}(\alpha, \gamma))$ are matrices corresponding to the aggregation and diversification operations utilizing the vector field $\nabla \phib^{(1)}(\alpha, \gamma)$
(see \eqref{eq:directedagg}).
Here $\mathbf{B}_{\{ \mathrm{av}, \mathrm{dx} \}}(\phib^{(1)}(\alpha, \gamma))_{ij}$
contains information of the diffusion distance between $v_i$ and $v_j$,
which describes the relative position of $v_i$ and $v_j$ on the graph.
Instead of regarding them as edge weights in DGN~\cite{beaini2021directional},
we use this node relative distance information to assist the learning of attention between nodes.
We propose to define the attention score at the $l^{th}$ layer between $v_i$ and $v_j$ before normalization as:
\be
\label{eq:dgat_attention}
r_{ij}^{(l)} (\alpha, \gamma) = \text{LeakyReLU}((\mathbf{a}^{(l)})^{\top} [\mathbf{W}_n^{(l)} \mathbf{h}_i^{(l-1)} || \mathbf{W}_n^{(l)} \mathbf{h}_j^{(l-1)} || \mathbf{W}_e^{(l)} \f^{(i,j)} (\alpha, \gamma) ]),
\ee
where 
$\mathbf{W}_n^{(l)}$ and $\mathbf{W}_e^{(l)}$ are learnable weights for node representation and edge features respectively.
We use $\alpha_{ij}^{(l)}(\alpha, \gamma)$ to denote the attention score after $\text{softmax}$ normalization.
The PD-GAT with multi-heads attention compute the new representation as: 
\be
\label{eq:dgat_multihead}
\h_i^{(l)} =\left( \bigg\Arrowvert^{M}_{m=1} \sigma \big( 
\sum_{j \in \{\mathcal{N}(i), i\}} \alpha_{ij}^{(l, m)} (\alpha, \gamma) \W^{(l, m)} \h_j^{(l-1)}
\big)   \right)\W^{(l)}
\ee
where $M$ is the number of heads.
Compared with GAT,
the additional number of parameters in PD-GAT introduced by \eqref{eq:dgat_attention} is negligible,
since $\f^{(i,j)} (\alpha, \gamma) \in \Rbb^2$.

\subsection{Topology-Guided Rewiring Mechanism for Long-range Diffusion}
\label{sec:rewire}
This subsection introduces a graph rewiring technique to further enhance the proposed parameterized diffusion on heterophilic graphs. Since diffusion distance is a criterion for the efficiency of message passing, we regard locally disconnected nodes with large diffusion distance as long-range neighbors. According to Theorem \ref{thm:gradient-v2}, large spectral distances also identify long-range neighbors. Therefore, the first non-trivial eigenvector $\phib^{(1)}(\alpha, \gamma)$ of $\mathbf{L}^{(\alpha, \gamma)}$ can be regarded as a one-dimensional embedding of nodes, and we refer to it as the (parameterized) spectral embedding. We refer to the node $v_{(\alpha, \gamma)}$ as the gradient node, whose corresponding element in $\phib^{(1)}(\alpha, \gamma)$ is the maximum, \ie{} $v_{(\alpha, \gamma)}$ is located at the rightmost position in the 1D spectral embedding.
In heterophilic graphs, our proposed rewiring strategy connects other nodes in the graph with the gradient node under a certain configuration of $\mathbf{L}^{(\alpha, \gamma)}$. The reasons for its effectiveness on heterophilic graph learning are provided below.

On heterophilic graphs, our proposed rewiring strategy aims to accelerate the diffusion process starting from a disconnected long-range neighbors $v_j$ to target node $v_i$. To find a proper candidate $v_j$ efficiently, we assume that: \textbf{(1)} $v_j$ is to the right of $v_i$ in the spectral embedding without loss of generality; \textbf{(2)} $v_i$ and the gradient node $v_{(\alpha, \gamma)}$, are locally disconnected in the original graph; \textbf{(3)} $\phib^{(1)}_j(\alpha, \gamma) - \phib^{(1)}_i(\alpha, \gamma) \geq \frac{1}{2} (\text{max}_m \phib^{(1)}_m(\alpha, \gamma) - \text{min}_n \phib^{(1)}_n(\alpha, \gamma))$ to reduce the number of long-range candidate neighbors of $v_i$. Based on Theorem \ref{thm:gradient-v2}, we know that the newly created connection by the rewiring strategy between $v_i$ and $v_{(\alpha, \gamma)}$ reduces the number of diffusion steps required for a message sent by $v_j$ to reach $v_i$. In other words, the diffusion from $v_j$ to $v_{(\alpha, \gamma)}$, and then from $v_{(\alpha, \gamma)}$ to $v_i$, is faster than the diffusion from $v_j$ to $v_i$ when $v_i$ and $v_{(\alpha, \gamma)}$ are disconnected. Therefore, the proposed rewiring benefit the long-range diffusion thus enhance the learning on heterophilic graphs.


\vspace{-0.3cm}
\section{Experiments}
\label{sec:experiments}
In this section, we evaluate the effectiveness of our proposed models on synthetic and real-world benchmark datasets. In Section \ref{sec:ablation}, we conduct ablation study to validate the effectiveness of each proposed component. In Section \ref{sec:synthetic}, we generate synthetic graphs to study how the performance of the proposed parameterized diffusion varies with the graph homophily levels. In Section \ref{sec:benchmark}, we compare the proposed architectures with baselines and state-of-the-art (SOTA) models on 7 real-world benchmark datasets and the results show that parameterized diffusion augmented GNNs outperform the SOTA models on 6 out of 7 node classification tasks. These results highlight the explainability of our parameterized diffusion with $\mathbf{L}^{(\alpha, \gamma)}$, and verify that the proposed GNNs indeed can adaptively capture global information by adjusting parameters in $\mathbf{L}^{(\alpha, \gamma)}$ for graphs with different levels of homophily.

\subsection{Ablation Study}
\label{sec:ablation}
\begin{table*}[ht!]
\resizebox{1\hsize}{!}{
  \begin{tabular}{c|cccccccccc}
  \toprule
    Method&
    $\mathbf{L}^{(\alpha, \gamma)}$&
    $\mathcal{PD}$&
    $\mathcal{R}(\mathcal{G})$&
    roman-empire&
    amazon-ratings&
    minesweeper&
    tolokers&
    questions&
    squirrel filtered&
    chameleon filtered
    \\
    \midrule
    \multirow{1}[0]{*}
    GAT&
    &
    &
    &
    80.87 $\pm$ 0.30&
    49.09 $\pm$ 0.63&
    92.01 $\pm$ 0.68&
    83.70 $\pm$ 0.47&
    77.43 $\pm$ 1.20&
    35.62 $\pm$ 2.06&
    39.21 $\pm$ 3.08
    \\

    \multirow{1}[0]{*}
    &
    &
    &
    $\checkmark$ &
    81.21 $\pm$ 0.71&
    47.27 $\pm$ 0.52&
    92.86 $\pm$ 0.53&
    84.28 $\pm$ 0.68&
    78.79 $\pm$ 1.01&
    40.45 $\pm$ 1.36&
    43.51 $\pm$ 5.06
    \\

        &
    $\checkmark$ &
        &
    $\checkmark$ &
    82.27 $\pm$ 0.60&
    47.69 $\pm$ 0.64&
    92.94 $\pm$ 0.55&
    84.59 $\pm$ 0.53&
    78.97 $\pm$ 1.04&
    41.49 $\pm$ 2.72&
    43.89 $\pm$ 4.67
    \\
    
    \midrule
    \multirow{1}[0]{*}
    &
    &
    $\checkmark$ &
    &
    82.50 $\pm$ 0.60&
    49.29 $\pm$ 0.57&
    92.07 $\pm$ 0.69&
    84.11 $\pm$ 0.41&
    77.82 $\pm$ 0.78&
    36.50 $\pm$ 1.37&
    41.84 $\pm$ 2.76
    \\

    \multirow{1}[0]{*}
    PD-GAT &
       &
    $\checkmark$ &
    $\checkmark$ &
    86.82 $\pm$ 0.60&
    47.81 $\pm$ 0.54&
    93.15 $\pm$ 0.80&
    84.39 $\pm$ 0.49&
    79.40 $\pm$ 0.79&
    41.48 $\pm$ 2.34&
    43.09 $\pm$ 4.13
    \\
    
    &
    $\checkmark$ &
    $\checkmark$ &
      &
    83.46 $\pm$ 0.39&
    \textbf{49.69 $\pm$ 0.51}&
    92.15 $\pm$ 0.71&
    84.11 $\pm$ 0.41&
    78.66 $\pm$ 0.97&
    37.83 $\pm$ 1.54&
    43.37 $\pm$ 3.01
    \\
    &
    $\checkmark$ &
    $\checkmark$ &
    $\checkmark$ &
    \textbf{87.27 $\pm$ 0.64}&
    48.03 $\pm$ 0.58&
    \textbf{93.27 $\pm$ 0.56}&
    \textbf{84.74 $\pm$ 0.59}&
    \textbf{79.55 $\pm$ 0.81}&
    \textbf{42.09 $\pm$ 2.65}&
    \textbf{44.16 $\pm$ 4.20}
    \\
    
    \midrule
    \midrule
    \multirow{1}[0]{*}
    &
    &
    &
    &
    73.69 $\pm$ 0.74& 
    48.70 $\pm$ 0.63&
    89.75 $\pm$ 0.52&
    83.64 $\pm$ 0.67&
    76.09 $\pm$ 1.27&
    39.47 $\pm$ 1.47&
    40.89 $\pm$ 4.12
    
    \\

    \multirow{1}[0]{*}
    GCN&
    &
    &
    $\checkmark$&
    74.50 $\pm$ 0.58&
    48.32 $\pm$ 0.54&
    89.93 $\pm$ 0.60&
    83.41 $\pm$ 0.93&
    77.24 $\pm$ 1.14&
    40.92 $\pm$ 1.49&
    43.60 $\pm$ 1.91
    
    \\

    \multirow{1}[0]{*}
    &
    $\checkmark$&
    &
    $\checkmark$&
    74.60 $\pm$ 0.65&
    48.45 $\pm$ 0.48&
    89.94 $\pm$ 0.61&
    83.46 $\pm$ 0.92&
    77.27 $\pm$ 1.14&
    41.12 $\pm$ 1.29&
    45.34 $\pm$ 4.53
    
    \\

    \midrule

    &
    &
    $\checkmark$&
    &
    73.97 $\pm$ 0.46&
    49.38 $\pm$ 0.80&
    91.60 $\pm$ 0.62&
    81.35 $\pm$ 0.66&
    74.80 $\pm$ 0.76&
    36.19 $\pm$ 2.45&
    41.65 $\pm$ 2.81
    
    \\

    \multirow{1}[0]{*}
    PD-GCN&
    $\checkmark$&
    $\checkmark$&
    &
    \textbf{78.05 $\pm$ 0.49}&
    \textbf{49.49 $\pm$ 0.67}&
    \textbf{91.60 $\pm$ 0.62}&
    \textbf{83.83 $\pm$ 0.86}&
    \textbf{78.04 $\pm$ 0.91}&
    \textbf{43.31 $\pm$ 1.92}&
    \textbf{46.67 $\pm$ 3.56}
    
    \\

  \bottomrule
  \end{tabular}
  }
\caption{Ablation study on real-world heterophily datasets proposed by \cite{platonov2023critical}.
Here, GCN and GAT are baseline models. A checkmark on $\mathbf{L}^{(\alpha, \gamma)}$ indicates the use of the parameterized normalized Laplacian (or the corresponding parameterized normalized adjacency $\mathbf{P}^{(\alpha, \gamma)}$) with optimally tuned $\alpha$ and $\gamma$, while an unchecked $\mathbf{L}^{(\alpha, \gamma)}$ denotes the use of the default $\mathbf{L}^{(1, 1)} = \mathbf{L}_{\rw}$ and $\mathbf{P}^{(1, 1)} = \mathbf{P}_{\rw}$. The term $\mathcal{PD}$ stands for ``parameterized diffusion", which refers to the weighted aggregation with $\mathbf{P}^{(\alpha, \gamma)}$ for PD-GCN and the incorporation of edge features defined by $\mathbf{L}^{(\alpha, \gamma)}$ in the graph attention for PD-GAT,
whereas unchecking $\mathcal{PD}$ means that the vanilla GCN or GAT is used.
Lastly, $\mathcal{R}(\mathcal{G})$ stands for the graph rewiring strategy.
Results for baselines with $\mathcal{R}(\mathcal{G})$ are obtained by applying baselines to the rewired graph, while results for baselines with both $\mathbf{L}^{(\alpha, \gamma)}$ and $\mathcal{R}(\mathcal{G})$ are from applying the rewiring technique with the optimally $\alpha$ and $\gamma$. Note that the rewiring technique is not applicable to PD-GCN, as PD-GCN assigns parameterized edge weights based on the original graph structure.
The best results are highlighted in \textbf{bold} format respectively for GCN and GAT-based architectures.}
\label{tab:heter_ablation}
\end{table*}

In this subsection, we conduct ablation study to investigate the effectiveness of (1) parameterized normalized Laplacian $\mathbf{L}^{(\alpha, \gamma)}$ and its corresponding adjacency matrix $\mathbf{P}^{(\alpha, \gamma)}$, (2) the topology-guided graph rewiring strategy and (3) the parameterized diffusion based GCN and GAT, referred to as PD-GCN and PD-GAT. The ablation results on 7 heterophily datasets are summarized in Table \ref{tab:heter_ablation}. Note that both the parameterized diffusion and the rewiring step are influenced by $\alpha$ and $\gamma$. The optimal $\alpha$ and $\gamma$ used in the ablation studies are provided in Appendix~\ref{app:Hyperparameters}.

From Table \ref{tab:heter_ablation}, 
we make the following observations:
(1) The rewiring strategy improves both baseline models and the proposed models;
(2) Comparing PD-GAT and GAT,
even without leveraging $\mathbf{L}^{(\alpha, \gamma)}$, \ie{} using only the default $\mathbf{L}_{\rw}$ with $\alpha=1$, $\gamma=1$, 
both the rewiring strategy and the parameterized diffusion can individually improve model performance; (3) PD-GCN is more sensitive to the choice of Laplacian parameters.
PD-GCN with the default parameters achieves performance gain over GCN, except on the \textit{questions} and \textit{squirrel-filtered}.
However, with optimal parameters,
PD-GCN even outperforms PD-GAT on these two datasets;
(4) After incorporating $\mathbf{L}^{(\alpha, \gamma)}$,
\ie{} perform a grid-search for the optimal $\alpha$, $\gamma$ for each graph to find the most suitable relative node positions, 
the effectiveness of 
the parameterized diffusion and the rewiring strategy 
is further enhanced; 
(5) Each component is indispensable for the success of PD-GCN and PD-GAT,
except for the \textit{amazon-ratings} dataset, 
where the rewiring does not provide a benefit\footnote{It can be explained by the results in~\cite{platonov2023critical}, which show that the GNNs yield negligible improvement over the graph-free models on \textit{amazon-ratings}.}.



\begin{figure*}[t]
\centering
 {  
 {
 \subfloat[]{
 \captionsetup{justification = centering}
 \includegraphics[width=0.5\textwidth]{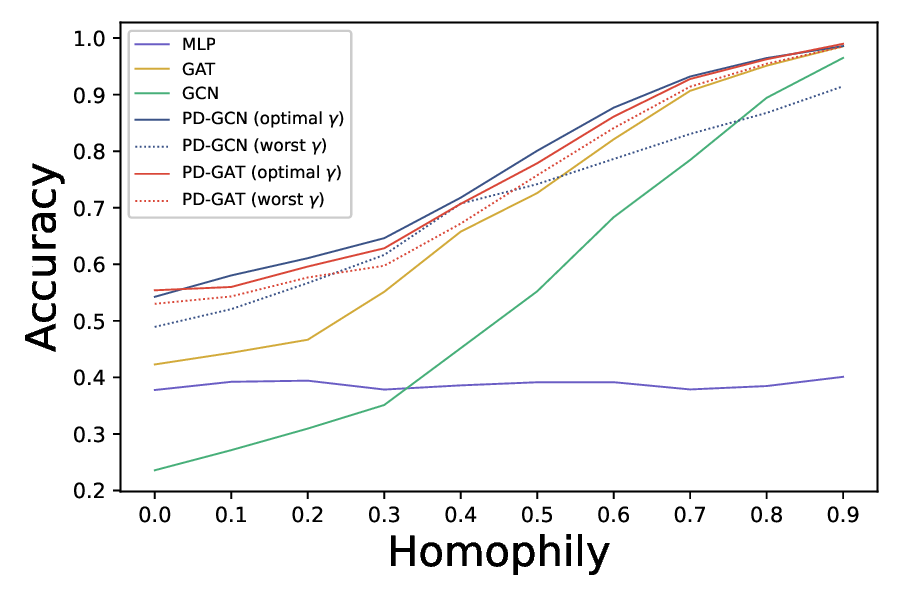}
 }
 \subfloat[]{
 \captionsetup{justification = centering}
 \includegraphics[width=0.5\textwidth]{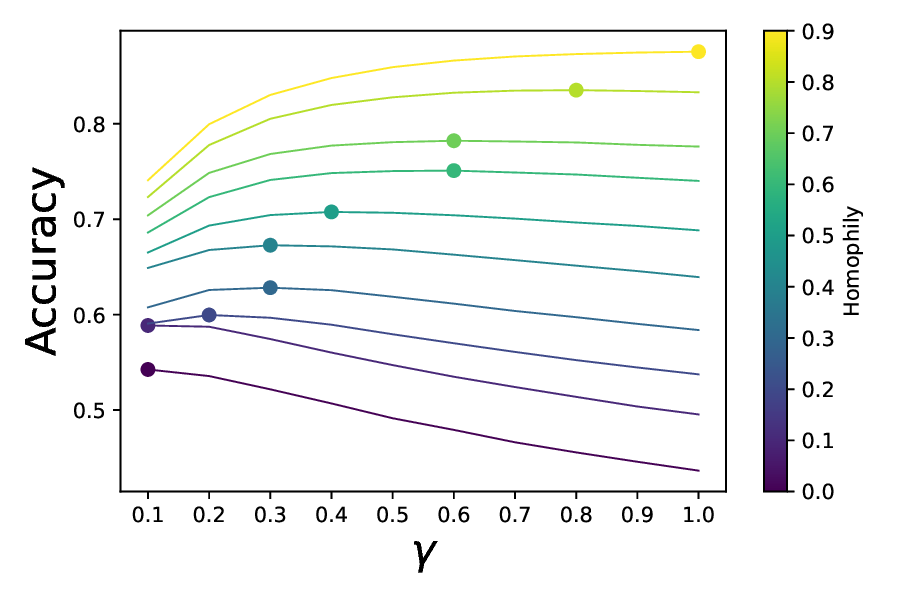}
 } 
 }
 }
 \caption{Experiments on synthetic graphs with varying levels of homophily. 
 The y-axis represents averaged test accuracy. 
 (a) Comparison of PD-GCN and PD-GAT with baseline models. 
 The x-axis denotes the graph homophily level, 
 where a larger value indicates a more homophilic graph.
 The solid lines for the proposed models represent performance with the optimal $\gamma$,
 while the dashed lines show performance with the worst $\gamma$.
 (b)
 Each line corresponds to synthetic graphs with a specific homophily level,
 illustrating the relationship between the performance of a one-layered PD-GCN and $\gamma$.
 The line color indicates the homophily coefficient,
 with blue representing low homophily and yellow indicating high homophily.
 The dot markers denote the optimal $\gamma$ for each homophily level.}
 \label{fig:syn_res}
\end{figure*}

\subsection{Synthetic Experiments}
\label{sec:synthetic}
In this subsection, we investigate the behavior of parameterized diffusion under different homophily levels, focusing on its performance and the optimal choice of $\gamma$. 
We follows~\cite{Karimi_2018, abu2019mixhop} to generate synthetic graphs characterized by a homophily coefficient $\mu \in \{0.0, 0.1, \ldots, 0.9 \}$, representing the chance that a node forms a connection to another node with the same label. We say the graph is more heterophilic for a smaller $\mu$. A detailed explanation about how the synthetic graph are generated is given in  Appendix \ref{appendix:syn-dataset}. We generate 5 synthetic graphs under each homophily coefficient $\mu$. Node features are sampled from overlapping multi-Gaussian distributions. And for each generated graph, nodes are randomly partitioned into train/validation/test sets with a ratio of 60\%/20\%/20\%. Each model (MLP, GAT, GCN, PD-GCN, PD-GAT) is trained under the same hyperparameter setting with the learning rate $0.01$, weight decay $0.001$ and dropout rate $0.1$. The number of layers is set to be 2 for each model. 
Each model use 64 hidden units and the attention-based model use 8 heads with 8 hidden states per head. For simplicity in comparison, the rewiring strategy is not applied to the proposed methods. For PD-GCN and PD-GAT, we fix $\alpha = 1.0$ and only vary $\gamma \in \{0.0, 0.1, \ldots, 0.9 \}$.

As shown in Figure~\ref{fig:syn_res} (a), both PD-GCN and PD-GAT outperform MLP, GCN and GAT across all homophily levels, particularly in heterophilic cases. PD-GCN with the optimal $\gamma$ achieves the best overall performance. However, the effectiveness of PD-GCN relies on a well-chosen $\gamma$, as its performance drops considerably with the worst $\gamma$, becoming worse than the baseline GNNs on more homophilic graphs. On the other hand, the performance of PD-GAT is more robust to $\gamma$. Even with the worst $\gamma$, PD-GAT still performs better results than the baselines. This difference arises because PD-GCN encodes the parameterized topology knowledge in the aggregation weights, whereas PD-GAT incorporates it into the edge features. As a result, the choice of parameter has a more direct impact on the performance of PD-GCN than PD-GAT.

Figure~\ref{fig:syn_res} (b) demonstrates the strong correlation between the parameter $\gamma$ and the model performance across different graph homophily levels. We find that: \textbf{(1)} As the graph becomes more heterophilic, the optimal $\gamma$ (the dot marker on each curve) for PD-GCN decreases, and vice versa. This suggests that the reduction of the diffusion distance between distant nodes, according to Theorem \ref{thm:gradient-v2}, is favored on heterophilic graphs. This also verifies that PD-GCN can indeed achieve flexible diffusion scopes by adjusting $\gamma$ and $\gamma$ is an interpretable parameter. \textbf{(2)} The performance changes smoothly with the value of $\gamma$. The curves corresponding to larger homophily levels are always higher than those with smaller homophily levels, which implies that, given a fixed $\gamma$, PD-GCN consistently performs better on more homophilic graphs. \textbf{(3)} Furthermore, the more heterophilic the graph is, the less satisfactory the performance with the default $\gamma = 1$. This phenomenon highlights that the limitations of the conventional Laplacian are significant in heterophilic cases, and the proposed new Laplacian effectively addresses this issue through parameterized diffusion.



\subsection{Experiments on Real-world Datasets}
\label{sec:benchmark}

In this subsection, we compare PD-GCN and PD-GAT with 6 baseline models\footnote{Baseline models adopt residual connections at each layer, following the implementation provided by~\cite{platonov2023critical}.}: GAT~\cite{velivckovic2017graph}, GAT-sep \footnote{Here "sep" means to concatenate the ego feature of node and the aggregated neighborhood information, which is the trick used in~\cite{platonov2023critical} and we follow this setting.},
GCN~\cite{kipf2016semi},
SAGE~\cite{hamilton2017inductive},
Graph Transformer (GT)~\cite{shi2020masked} and
GT-sep,
and 11 heterophily-specific SOTA models:
H$_2$GCN~\cite{zhu2020beyond},
CPGNN~\cite{zhu2021graph},
GPR-GNN~\cite{chien2020adaptive},
FSGNN~\cite{maurya2022simplifying},
GloGNN~\cite{li2022finding},
FAGCN~\cite{bo2021beyond},
GBK-GNN~\cite{du2022gbk},
JacobiCov~\cite{wang2022powerful},
BernNet~\cite{he2021bernnet},
LINKX~\cite{lim2021large} and
APPNP~\cite{gasteiger2018predict}.
The comparison also contains GNNs with novel aggregation including
DGN~\cite{beaini2021directional},
$\omega$GCN~\cite{eliasof2023improving}
and PGSO-GCN~\cite{dasoulas2021learning}.
The experiments are conducted on 7 recently proposed heterophily benchmark datasets~\cite{platonov2023critical}:
\textit{roman-empire},
\textit{amazon-ratings},
\textit{minesweeper},
\textit{tolokers},
\textit{questions},
\textit{Chameleon-filtered} and \textit{Squirrel-filtered}\footnote{The overall statistics of these real-word datasets are given in Appendix \ref{appendix:real-dataset}. And additional results on homophilic datasets are provided in Appendix \ref{app:homophily}}. 

\begin{table*}[t]
 \resizebox{1\hsize}{!}
 {
  \begin{tabular}{ccccccccccc}
  \toprule

    &
    roman-empire&
    amazon-ratings&
    minesweeper&
    tolokers&
    questions&
    squirrel-filtered&
    chameleon-filtered
    \\

    \midrule
    \multirow{1}[0]{*}
    GCN&
    73.69 $\pm$ 0.74& 
    48.70 $\pm$ 0.63&
    89.75 $\pm$ 0.52&
    83.64 $\pm$ 0.67&
    76.09 $\pm$ 1.27&
    39.47 $\pm$ 1.47&
    40.89 $\pm$ 4.12
    \\
    
    \multirow{1}[0]{*}
    SAGE&
    85.74 $\pm$ 0.67&
    \red{53.63 $\pm$ 0.39}&
    93.51 $\pm$ 0.57&
    82.43 $\pm$ 0.44&
    76.44 $\pm$ 0.62&
    36.09 $\pm$ 1.99&
    37.77 $\pm$ 4.14
    \\

    \multirow{1}[0]{*}
    GAT&
    80.87 $\pm$ 0.30&
    49.09 $\pm$ 0.63&
    92.01 $\pm$ 0.68&
    83.70 $\pm$ 0.47&
    77.43 $\pm$ 1.20&
    35.62 $\pm$ 2.06&
    39.21 $\pm$ 3.08
    \\

    \multirow{1}[0]{*}
    GAT-sep&
    \blue{88.75 $\pm$ 0.41}&
    \textcolor{violet}{52.70 $\pm$ 0.62}&
    \blue{93.91 $\pm$ 0.35}&
    83.78 $\pm$ 0.43&
    76.79 $\pm$ 0.71&
    35.46 $\pm$ 3.10&
    39.26 $\pm$ 2.50
    \\

    \multirow{1}[0]{*}
    GT&
    86.51 $\pm$ 0.73&
    51.17 $\pm$ 0.66&
    91.85 $\pm$ 0.76&
    83.23 $\pm$ 0.64&
    77.95 $\pm$ 0.68&
    36.30 $\pm$ 1.98&
    38.87 $\pm$ 3.66
    \\
    
    \multirow{1}[0]{*}
    GT-sep&
    87.32 $\pm$ 0.39&
    52.18 $\pm$ 0.80&
    92.29 $\pm$ 0.47&
    82.52 $\pm$ 0.92&
    78.05 $\pm$ 0.93&
    36.66 $\pm$ 1.63&
    40.31 $\pm$ 3.01
    \\

    \midrule
    \multirow{1}[0]{*}
    H$_2$GCN&
    60.11 $\pm$ 0.52&
    36.47 $\pm$ 0.23&
    89.71 $\pm$ 0.31&
    73.35 $\pm$ 1.01&
    63.59 $\pm$ 1.46&
    35.10 $\pm$ 1.15&
    26.75 $\pm$ 3.64
    \\
    
    \multirow{1}[0]{*}
    CPGNN&
    63.96 $\pm$ 0.62&
    39.79 $\pm$ 0.77&
    52.03 $\pm$ 5.46&
    73.36 $\pm$ 1.01&
    65.96 $\pm$ 1.95&
    30.04 $\pm$ 2.03&
    33.00 $\pm$ 3.15
    \\

    \multirow{1}[0]{*}
    GPR-GNN&
    64.85 $\pm$ 0.27&
    44.88 $\pm$ 0.34&
    86.24 $\pm$ 0.61&
    72.94 $\pm$ 0.97&
    55.48 $\pm$ 0.91&
    38.95 $\pm$ 1.99&
    39.93 $\pm$ 3.30
    \\

    \multirow{1}[0]{*}
    FSGNN&
    79.92 $\pm$ 0.56&
    \blue{52.74 $\pm$ 0.83}&
    90.08 $\pm$ 0.70&
    82.76 $\pm$ 0.61&
    \textcolor{violet}{78.86 $\pm$ 0.92}&
    35.92 $\pm$ 1.32&
    40.61 $\pm$ 2.97
    \\

    \multirow{1}[0]{*}
    GloGNN&
    59.63 $\pm$ 0.69&
    36.89 $\pm$ 0.14&
    51.08 $\pm$ 1.23&
    73.39 $\pm$ 1.17&
    65.74 $\pm$ 1.19&
    35.11 $\pm$ 1.24&
    25.90 $\pm$ 3.58
    \\

    \multirow{1}[0]{*}
    FAGCN&
    65.22 $\pm$ 0.56&
    44.12 $\pm$ 0.30&
    88.17 $\pm$ 0.73&
    77.75 $\pm$ 1.05&
    77.24 $\pm$ 1.26&
    \textcolor{violet}{41.08 $\pm$ 2.27}&
    41.90 $\pm$ 2.72
    \\

    \multirow{1}[0]{*}
    GBK-GNN&
    74.57 $\pm$ 0.47&
    45.98 $\pm$ 0.71&
    90.85 $\pm$ 0.58&
    81.01 $\pm$ 0.67&
    74.47 $\pm$ 0.86&
    35.51 $\pm$ 1.65&
    39.61 $\pm$ 2.60
    \\

    \multirow{1}[0]{*}
    JacobiCov&
    71.14 $\pm$ 0.42&
    43.55 $\pm$ 0.48&
    89.66 $\pm$ 0.40&
    68.66 $\pm$ 0.65&
    73.88 $\pm$ 1.16&
    29.71 $\pm$ 1.66&
    39.00 $\pm$ 4.20
    \\

    \multirow{1}[0]{*}
    BernNet&
    65.56 $\pm$ 1.34&
    44.64 $\pm$ 0.56&
    77.99 $\pm$ 0.95&
    77.00 $\pm$ 0.65&
    70.43 $\pm$ 1.38&
    41.18 $\pm$ 1.77&
    40.90 $\pm$ 4.06
    \\

    \multirow{1}[0]{*}
    LINKX&
    56.15 $\pm$ 0.93&
    52.66 $\pm$ 0.64&
    56.78 $\pm$ 2.47&
    81.15 $\pm$ 1.23&
    71.96 $\pm$ 2.07&
    40.10 $\pm$ 2.21&
    42.34 $\pm$ 4.13
    \\

    \multirow{1}[0]{*}
    APPNP&
    65.87 $\pm$ 0.53&
    46.02 $\pm$ 0.73&
    69.62 $\pm$ 2.11&
    76.98 $\pm$ 1.03&
    64.77 $\pm$ 1.32&
    35.12 $\pm$ 1.12&
    37.50 $\pm$ 3.69
    \\
    
    \midrule
    \multirow{1}[0]{*}
    DGN&
    83.12 $\pm$ 0.47&
    47.65 $\pm$ 0.71&
    90.64 $\pm$ 0.61&
    OOM&
    OOM&
    38.56 $\pm$ 1.84&
    41.24 $\pm$ 3.62
    \\

    \multirow{1}[0]{*}
    \text{$\omega$}GCN &
    74.79 $\pm$ 0.46&
    51.79 $\pm$ 0.74&
    90.71 $\pm$ 0.67&
    80.96 $\pm$ 1.04&
    71.10 $\pm$ 1.45&
    35.65 $\pm$ 2.06&
    40.93 $\pm$ 3.48&
    \\

    \multirow{1}[0]{*}
    PGSO-GCN&
    OOM&
    OOM&
    OOM&
    OOM&
    OOM&
    41.06 $\pm$ 3.05&
    \textcolor{violet}{43.91 $\pm$ 3.04}
    \\
    \midrule


    \multirow{1}[0]{*}
    PD-GCN&
    78.05 $\pm$ 0.49&
    49.49 $\pm$ 0.67&
    91.60 $\pm$ 0.62&
    83.83 $\pm$ 0.86&
    78.04 $\pm$ 0.91&
    \red{43.31 $\pm$ 1.92}&
    \red{46.67 $\pm$ 3.56}
    \\


    \multirow{1}[0]{*}
    PD-GAT&
    83.46 $\pm$ 0.39&
    49.69 $\pm$ 0.51&
    92.15 $\pm$ 0.71&
    \textcolor{violet}{84.11 $\pm$ 0.41}&
    78.66 $\pm$ 0.97&
    37.83 $\pm$ 1.54&
    43.37 $\pm$ 3.01
    \\
    
    \multirow{1}[0]{*}
    PD-GAT ($\mathcal{R}(\mathcal{G})$)&
    87.27 $\pm$ 0.64&
    48.03 $\pm$ 0.58&
    93.27 $\pm$ 0.56&
    \blue{84.74 $\pm$ 0.59}&
    \red{79.55 $\pm$ 0.81}&
    \blue{42.09 $\pm$ 2.65}&
    \blue{44.16 $\pm$ 4.20}
    \\

    \multirow{1}[0]{*}
    PD-GAT-sep&
    \textcolor{violet}{88.46 $\pm$ 0.58}&
    52.57 $\pm$ 0.95&
    \textcolor{violet}{93.81 $\pm$ 0.42}&
    84.04 $\pm$ 0.51&
    77.29 $\pm$ 0.73&
    37.01 $\pm$ 2.97&
    41.40 $\pm$ 4.74
    \\

    \multirow{1}[0]{*}
    PD-GAT-sep ($\mathcal{R}(\mathcal{G})$)&
    \red{89.23 $\pm$ 0.56} &
    50.96 $\pm$ 0.43 &
    \red{94.03 $\pm$ 0.45}&
    \red{84.83 $\pm$ 0.40}&
    \blue{78.88 $\pm$ 0.94}&
    39.69 $\pm$ 2.28&
    41.15 $\pm$ 4.66
    \\
  \bottomrule
  \end{tabular}
  }
  \caption{Experiment results on heterophily datasets proposed by~\cite{platonov2023critical}.
  Values stand for mean and standard deviation of evaluation metrics on the test datasets.
  Here
  \textit{roman-empire}, \textit{amazon-ratings}, \textit{squirrel-filtered} and \textit{chameleon-filtered} use accuracy for evaluation, while \textit{minsweeper}, \textit{tolokers} and \textit{questions} use ROC AUC. The "sep" refers to the trick proposed in~\cite{zhu2020beyond} which concatenates node's and the mean of neighbours' embedding in each aggregation step, instead of adding them together. For fair comparison, we use the same experimental setup in~\cite{platonov2023critical} for proposed models. Results for heterophily GNNs, except for BernNet, LINKX and APPNP, are reported by~\cite{platonov2023critical}. The top three results are highlighted in \red{red}, \blue{blue}, and \textcolor{violet}{violet}, respectively.}
  \label{table:heterophily_new}
\end{table*}

\paragraph{Experimental Setups} For the 7 new heterophily datasets, we evaluate PD-GCN and PD-GAT and their variants under the same experiment settings for baseline models with fixed splits used in~\cite{platonov2023critical},
with learning rate of $3 \cdot 10^{-5}$, 
weight decay of 0, 
dropout rate of 0.2, 
hidden dimension of 512 and attention head of 8, 
and number of layers from 1 to 5. Following~\cite{platonov2023critical}, the proposed models are also trained for 1000 steps with Adam optimizer and select the best step based on the performance on the validation set.
For other GNNs,
we perform a grid search for the learning rate, weight decay and dropout rate, and model-specific hyperparameters, with details provided in Appendix \ref{app:Hyperparameters}. 

The results are summarized in Table \ref{table:heterophily_new}, where GNN baselines are organized in the top block, heterophily-specific GNNs are placed in the second block,
models with novel aggregation operators are in the third block, and our proposed methods are listed in the bottom block.

\paragraph{Results and Comparisons} \textbf{(1)} In most cases, the proposed parameterized diffusion augmented GNNs obtain significant performance improvement against the baseline models, \ie{} PD-GCN, PD-GAT, PD-GAT-sep outperform GCN, GAT and GAT-sep, respectively. \textbf{(2)} The proposed models consistently surpass SOTA GNNs specifically designed to address heterophily, as well as GNNs utilizing various aggregation paradigms. The "sep" trick further enhances PD-GAT's performance on \textit{roman-empire}, \textit{minesweeper} and \textit{tolokers}. The ``sep" trick allows the aggregation step to assign negative weights to the propagated message, enabling node-wise diversification, which has been shown to be useful for heterophily data~\cite{luan2021heterophily,luan2022complete}. \textbf{(3)} In addition, the superior performance of PD-GAT over GT-based methods indicates that finding the useful graph for message propagation with sparse attention is more effective than considering message passing between all pairs of nodes. \textbf{(4)} The topology-guided rewiring mechanism is effective for GAT and GAT-sep on most real-world benchmark datasets.

\vspace{-0.3cm}
\section{Conclusion}
In this paper, we address the limitations of GNNs in capturing long-range and global topology information, particularly in heterophilic graphs, by proposing a novel class of parameterized normalized Laplacian matrices. The new Laplacian provides greater flexibility in controlling the diffusion distance between nodes, enabling adaptive diffusion scopes to accommodate varying levels of graph homophily. Then, we prove that the order-preserving relationship between the diffusion distance and spectral distance. With this result and the new Laplacian, we propose two models with flexible diffusion scopes, PD-GCN and PD-GAT, along with a topology-guided rewiring strategy that further enhances performance. The effectiveness of the proposed methods is justified both theoretically and empirically.


\clearpage

\bibliography{reference}
\bibliographystyle{unsrtnat}
\clearpage
\appendix

\section{Homophily Metrics}
\label{app:metric}
Here we review some commonly used metrics to measure homophily~\cite{luan2024heterophilic, zheng2024missing}.
We denote $\z \in \Rbb^N$ as labels of nodes,
and $\mathbf{Z} \in \Rbb^{N\times C}$ as the one-hot encoding of labels,
where $C$ is the number of classes.
Edge homophily $H_{\mathrm{edge}}$ and node homophily $H_{\mathrm{node}}$ are defined as follows:
\begin{align}
H_{\mathrm{edge}} = \frac{\{ e_{ij}| e_{ij} \in \mathcal{E}, \z_i = \z_j \}}{|\mathcal{E}|}, \: H_{\mathrm{node}} = \frac{1}{|\mathcal{V}|} \sum_{u \in \mathcal{V}}\frac{|y_u = y_v: v \in \mathcal{N}(v)| }{|\mathcal{N}(u)|}.
\end{align}
Adjusted edge homophily $H_{\mathrm{edge}}^{*}$
considers classes imbalance and is defined as~\cite{platonov2023critical} :
\begin{align}
H_{\mathrm{edge}}^{*} = \frac{H_{\mathrm{edge}} - \sum_c p^2(c)}{1- \sum_c p^2(c)}.
\end{align}
Here $p(c) = \sum_{i: \z_i=c} \D_{ii} / (2 |\mathcal{E}|)$, $c=1:C$, defines the degree-weighted distribution of class labels.
The class homophily is also proposed to take class imbalance into account~\cite{lim2021new}:
\begin{align}
&H_{\mathrm{class}} = \frac{1}{C-1}\sum_{c}\left[h_c - \frac{|\{v_i|\z_i = c\}|}{N} \right]_{+}\\   
&h_c = \frac{\sum_{v_i: \z_i = c} |\{e_{ij}|  e_{ij} \in \mathcal{E}, \z_i = \z_j \}|}{\sum_{v_i: \z_i = c} \D_{ii}}.
\end{align}
Label informativeness,
which indicates the amount of information a neighbor's label provides about node's label,
is defined as follows~\cite{platonov2023critical}:
\begin{align}
LI = 2 - \frac{\sum_{c_1, c_2} p(c_1, c_2) \log p(c_1, c_2)}{\sum_c p(c) \log p(c)},  
\end{align}
where $p(c_1, c_2) = |\{e_{ij}|e_{ij} \in \mathcal{E}, \z_i = c_1, \z_j = c_2 \}|/(2|\mathcal{E}|)$.
The aggregation homophily $H_{\mathrm{agg}}^{\mathrm{M}}(\mathcal{G})$ measures the proportion of nodes that assign greater average weights to intra-class nodes than inter-class nodes.
It is defined as follows~\cite{luan2022revisiting}:
\begin{align}
H_{\mathrm{agg}}^{\mathrm{M}}(\mathcal{G}) = \frac{1}{|\mathcal{V}|} \big| \{ &v_i| 
\mathrm{Mean}_{j} (\{S(\hat{\mathbf{A}}, \Z)_{ij}|\z_i=\z_j \} ) \\
& \geq \mathrm{Mean}_{j}(\{S(\hat{\mathbf{A}}, \Z)_{ij}|\z_i\neq \z_j \} )\} \big| \nonumber, 
\end{align}
where $S(\hat{\mathbf{A}}, \Z) = \hat{\mathbf{A}}\Z (\hat{\mathbf{A}}\Z)^\top$ defines the post-aggregation node similarity,
with $\hat{\mathbf{A}} = \mathbf{A} + \mathbf{I}$,
and $\mathrm{Mean}_{j}(\{\cdot\})$ takes the average over node $v_j$ of a given multiset of values.
Under the homophily metrics mentioned above,
a smaller value  
indicates a higher degree of heterophily.
While $H_{\mathrm{edge}}^{*}$ can assume negative values,
the other metrics fall within the range $[0, 1]$.

 



\section{Proof of Theorem}
\subsection{Proof of theorem \ref{thm:parameterized-matrix}}
\label{appendix:proof_parameterized_matrix}
\begin{proof}
 By Definition \ref{def:aug-lap}, we have
\begin{align*}
& \P^{(\alpha, \gamma)} = \I - \mathbf{L}^{(\alpha,\gamma)}
    = \I - \gamma [\gamma \D + (1-\gamma)\I]^{-\alpha} \mathbf{L}[\gamma\D + (1-\gamma)\I]^{\alpha-1} \\
& =[\gamma \D + (1-\gamma)\I]^{-\alpha}
  [\gamma \D + (1-\gamma)\I - \gamma \mathbf{L}]
 [\gamma \D + (1-\gamma)\I]^{\alpha-1} \\
& =[\gamma \D + (1-\gamma)\I]^{-\alpha}
  [\gamma \A + (1-\gamma)\I]
 [\gamma \D + (1-\gamma)\I]^{\alpha-1}
\end{align*}
It is easy to see that all elements in $\P^{(\alpha,\gamma)}$ are non-negative. 
Since $\A\1=\D\1$, we have
$$
\P^{(1,\gamma)} \1 = (\gamma\D + (1-\gamma)\I)^{-1}
        \left(\gamma\A+(1-\gamma)\I\right) \1 = \1,
$$
completing the proof.
\end{proof}

\subsection{Proof of theorem \ref{thm:eigenplaplaican}}
\label{appendix:proof_eigenplaplaican}
\begin{proof}
For any nonzero $\x\in \Rbb^N$, 
write $\y:=[\gamma \D + (1-\gamma)\I]^{-1/2}\x$.
Then we have
\begin{align*}
& \frac{\x^\top \mathbf{L}^{(1/2, \gamma)}\x}{\x^\top\x} = \frac{\x^\top\gamma[\gamma \D + (1-\gamma)\I]^{-1/2} \mathbf{L}
[\gamma \D + (1-\gamma)\I]^{-1/2}\x}{\x^\top\x} \\
& = \frac{\gamma\y^\top \mathbf{L} \y}{\y^\top [\gamma \D + (1-\gamma)\I] \y} =\frac{\frac{\gamma}{2} \sum_{ij} a_{ij}(y_i-y_j)^2}
{\gamma \sum_{ij} a_{ij}y_i^2 + (1 - \gamma) \sum_{i}y_i^2}
\end{align*}
By the Rayleigh quotient theorem,  
\be \label{eq:rayleighr}
\lambda^{(0)}(\gamma)
=\min_{\y \neq \0}  \frac{ \frac{\gamma}{2}\sum_{ij} a_{ij}(y_i-y_j)^2}
{\gamma\sum_{ij} a_{ij}y_i^2+ (1 - \gamma) \sum_{i}y_i^2} = 0,
\ee
where the minimum is reached when $\y$ is a multiple of $\1$
and
\begin{align*}
\lambda^{(N-1)}(\gamma)
& = \max_{\y \neq \0} \frac{ \frac{\gamma}{2}\sum_{ij} a_{ij}(y_i-y_j)^2}
{\gamma\sum_{ij} a_{ij}y_i^2+ (1 - \gamma)\sum_{i}y_i^2}  \\
&  \leq  \max_{\y \neq \0} \frac{ \sum_{ij} a_{ij}  (y_i^2 +y_j^2)}
 {\sum_{ij} a_{ij}y_i^2}  
 \leq 2,
\end{align*}
leading to Eq.~\eqref{eq:glevrange}.
The proof of showing $\lambda^{(1)}(\gamma)\neq 0$ if and only if ${\cal G}$ is connected is similar to~\cite{Belkin2003}, thus we omit the details here.

By the Courant-Fischer min-max theorem,  
for $\gamma\neq 0$,
\begin{align*}
    \lambda^{(i)}(\gamma)
& = \min_{\{S:\text{dim}(S)=i+1\}} \max_{\{\x:\0\neq \x\in S\}} 
\frac{\x^\top \mathbf{L}^{(1/2, \gamma)}\x}{\x^\top\x} \\
& = \min_{\{S:\text{dim}(S)=i+1\}}\max_{\{\y:\0\neq \y\in S\}} 
\frac{\y^\top\mathbf{L}\y}{\y^\top[\D-\I+ (1/\gamma)\I]\y}.
\end{align*}
It is obvious that the Rayleigh quotient
$\frac{\y^\top\mathbf{L}\y}{\y^\top [\D-\I+ (1/\gamma)\I] \y}$
is strictly increasing with respect to $\gamma\in (0,1]$ if $\mathbf{L}\y\neq 0$,
\ie{} $\y$ not a multiple of $\1$. 
Note that $\lambda^{(0)}(\gamma)=0$ and it is reached when 
$\mathbf{L}\y=\0$, or  equivalently $\y$ is a multiple of $\1$.
Thus, $\lambda^{(i)}(\gamma)$ is strictly increasing with respect to $\gamma$
for $i=1:N-1$.

From the eigendecomposition of the symmetric $\mathbf{L}^{(1/2,\gamma)}$ in ~\eqref{eq:gled},
we can find the eigendecomposition of $\mathbf{L}^{(\alpha,\gamma)}$
as follows:
\begin{align*}
&\mathbf{L}^{(\alpha,\gamma)} = [\gamma\D+(1-\gamma)\I]^{1/2-\alpha}\mathbf{L}^{(1/2, \gamma)} [\gamma\D+(1-\gamma)\I]^{\alpha -1/2} \\
& = [\gamma\D+(1-\gamma)\I]^{1/2-\alpha}(\U \bLambda^{(\gamma)} \U^\top)
[\gamma\D+(1-\gamma)\I]^{\alpha -1/2}\\
& = \left([\gamma\D+(1-\gamma)\I]^{1/2-\alpha}\U\right) \bLambda^{(\gamma)}  
\left([\gamma\D+(1-\gamma)\I]^{1/2-\alpha}\U\right)^{-1}
\end{align*}
Thus,  $\lambda^{(i)} (\gamma)$ is also 
an eigenvalue of $\mathbf{L}^{(\alpha,\gamma)}$ for $i=0:N-1$,
and the $i$-th column of $[\gamma\D+(1-\gamma)\I]^{1/2-\alpha}\U$ is
a corresponding eigenvector.
\end{proof}

\subsection{Proof of Theorem \ref{thm:gradient-v2}}
\label{appendix:gradient-v2}
\begin{proof}
The proof is similar to the proof of~\cite{beaini2021directional}.
By~\cite{Coifman2006Diffusion}, 
the diffusion distance at time $t$ between node $v_i$ and $v_j$ can be expressed as:
    \be
    \label{eq:pf-dt2}
       d_t(v_i, v_j) = \left( \sum_{k=1}^{n-1} e^{-2t \lambda^{(k)}(\gamma)}(\phi_i^{(k)}(\gamma) - \phi_j^{(k)}(\gamma))^2 \right)^\frac{1}{2},
    \ee
    where 
    $
    \lambda^{(1)}(\gamma) \leq \lambda^{(2)}(\gamma) \leq \dots \leq \lambda^{(n-1)}(\gamma)
    $
    are eigenvalues of $\mathbf{L}^{(1, \gamma)}$, and $\{\phib^{(1)}(\gamma), \phib^{(2)}(\gamma), \dots, \phib^{(n-1)}(\gamma)\}$ are the corresponding eigenvectors. We omit the zero $\lambda^{(0)}(\gamma)$.
    The inequality $d_t(v_m, v_j) < d_t(v_i, v_j)$ is then equivalent as 
    \be
        \begin{aligned}
        \left( \sum_{k=1}^{n-1} e^{-2t \lambda^{(k)}(\gamma)}(\phi_m^{(k)}(\gamma) - \phi_j^{(k)}(\gamma))^2 \right)^\frac{1}{2}&\\
        < \left( \sum_{k=1}^{n-1} e^{-2t \lambda^{(k)}(\gamma)}(\phi_i^{(k)}(\gamma) - \phi_j^{(k)}(\gamma))^2
        \right)^\frac{1}{2}.
        \end{aligned}
    \ee
    We can take out $\lambda^{(1)}(\gamma)$ and $\phib^{(1)}(\gamma)$ and rearrange the above inequality as:
    \be
        \begin{aligned}
            \label{eq:pf-ineq}
            &\sum_{k=2}^{n-1} e^{-2t \lambda^{(k)}(\gamma)}
            \left(
            (\phi_m^{(k)}(\gamma) - \phi_j^{(k)}(\gamma))^2 - 
            (\phi_i^{(k)}(\gamma) - \phi_j^{(k)}(\gamma))^2 
            \right)\\
            & < e^{-2t \lambda^{(1)}(\gamma)}\left(
            (\phi_i^{(1)}(\gamma) - \phi_j^{(1)}(\gamma))^2 - 
            (\phi_m^{(1)}(\gamma) - \phi_j^{(1)}(\gamma))^2 
            \right).
        \end{aligned}
    \ee
    The left-hand side of Eq.~\eqref{eq:pf-ineq} 
    has an upper bound:
    \be
    \begin{aligned}
         \sum_{k=2}^{n-1} e^{-2t \lambda^{(k)}(\gamma)}
        \left|
        (\phi_m^{(k)}(\gamma) - \phi_j^{(k)}(\gamma))^2 - 
        (\phi_i^{(k)}(\gamma) - \phi_j^{(k)}(\gamma))^2 
        \right|\\
        \leq
        e^{-2t \lambda^{(2)}(\gamma)}
        \sum_{k=2}^{n-1} 
        \left|
        (\phi_m^{(k)}(\gamma) - \phi_j^{(k)}(\gamma))^2 - 
        (\phi_i^{(k)}(\gamma) - \phi_j^{(k)}(\gamma))^2 
        \right|.
    \end{aligned}
    \ee
    Then Eq.~\eqref{eq:pf-ineq} holds if:
    \be
        \begin{aligned}
            \label{eq:pf-ineq2}
             e^{-2t \lambda^{(2)}(\gamma)}
            \sum_{k=2}^{n-1} 
            \left|
            (\phi_m^{(k)}(\gamma) - \phi_j^{(k)}(\gamma))^2 - 
            (\phi_i^{(k)}(\gamma) - \phi_j^{(k)}(\gamma))^2 
            \right|\\
            \leq
            e^{-2t \lambda^{(1)}(\gamma)}\left(
            (\phi_i^{(1)}(\gamma) - \phi_j^{(1)}(\gamma))^2 - 
            (\phi_m^{(1)}(\gamma) - \phi_j^{(1)}(\gamma))^2 
            \right),
        \end{aligned}
    \ee
    which is equivalent to 
    
 \be
    \begin{aligned}
        \label{eq:pf-ineq3}
        &\log\left(
        \frac
        {(\phi_i^{(1)}(\gamma) - \phi_j^{(1)}(\gamma))^2 - 
        (\phi_m^{(1)}(\gamma) - \phi_j^{(1)}(\gamma))^2}
        {\sum_{k=2}^{n-1} 
        \left|
        (\phi_m^{(k)}(\gamma) - \phi_j^{(k)}(\gamma))^2 - 
        (\phi_i^{(k)}(\gamma) - \phi_j^{(k)}(\gamma))^2 
        \right|}
        \right)\\
        & \times \frac{1}{2(\lambda^{(1)}(\gamma) - \lambda^{(2)}(\gamma))} < t.
    \end{aligned}
    \ee      
    Let the constant $C$ be the left-hand side of Eq.~\eqref{eq:pf-ineq3}, 
    then if we take $t \geq \left \lfloor C \right \rfloor + 1$, 
    we have $d_t(v_m, v_j) < d_t(v_i, v_j)$.    
    Note that $C$ exits if
    \be
        \frac{(\phi_i^{(1)}(\gamma) - \phi_j^{(1)}(\gamma))^2 - 
        (\phi_m^{(1)}(\gamma) - \phi_j^{(1)}(\gamma))^2}
        {\sum_{k=2}^{n-1} 
        \left|
        (\phi_m^{(k)}(\gamma) - \phi_j^{(k)}(\gamma))^2 - 
        (\phi_i^{(k)}(\gamma) - \phi_j^{(k)}(\gamma))^2 
        \right|} > 0,
    \ee
    which is satisfied since we assume
    $| \phi_i^{(1)}(\gamma) - \phi_j^{(1)}(\gamma) | > | \phi_m^{(1)}(\gamma) - \phi_j^{(1)}(\gamma)|$.
    The original theorem~\cite{beaini2021directional} is only based on $\phib^{(1)}$ and does not assume that $v_m$ must satisfy $| \phi_i^{(1)} - \phi_j^{(1)} | > | \phi_m^{(1)} - \phi_j^{(1)}|$,
    which is necessary for the existence of $C$.
    In addition, the original theorem~\cite{beaini2021directional} assumes that $v_m$ is obtained by taking a gradient step from $v_i$, \ie{} $\phi_m - \phi_i = \max_{j:v_j \in \calN(v_i)} (\phi_j - \phi_i)$,
    while this property is not needed for the proof.
    Therefore,
    Theorem \ref{thm:gradient-v2} both extends and overcomes the shortcomings of~\cite{beaini2021directional}.
\end{proof}

\section{Datasets}
\subsection{Real-world Datasets}
\label{appendix:real-dataset}
The overall statistics of the real-world datasets are presented in Table \ref{table:chars} and
Table \ref{table:heter_metric} provides their heterophily levels calculated using various homophily metrics.
\begin{table}[htbp]
  \centering  
  \begin{tabular}{lcccccc}
  \toprule
  &
  \#Nodes&
  \#Edges&
  \#Features&
  \#Classes&
  Metric
  \\

  \midrule






  cora&
  2,708&
  5,278&
  1,433&
  7&
  ACC
  \\

  citeseer&
  3,327&
  4,552&
  3,703&
  6&
  ACC
  \\

  pubmed&
  19,717&
  44,324&
  500&
  3&
  ACC
  \\

  roman-empire&
  22,662&
  32,927&
  300&
  18&
  ACC
  \\

  amazon-ratings&
  24,492&
  93,050&
  300&
  5&
  ACC
  \\

  minesweeper&
  10,000&
  39,402&
  7&
  2&
  ROC AUC
  \\

  tolokers&
  11,758&
  519,000&
  10&
  2&
  ROC AUC
  \\

  questions&
  48,921&
  153,540&
  301&
  2&
  ROC AUC
  \\

  squirrel-filtered&
  2,223&
  46,998&
  2,089&
  5&
  ACC
  \\

  chameleon-filtered&
  890&
  8,854&
  2,325&
  5&
  ACC
  \\

  \bottomrule
  \end{tabular}
    \caption{
    Statistics of the benchmark dataset. Following pre-processing, the graph has been transformed into an undirected and simple form, without self-loops or multiple edges.}
\end{table}
\label{table:chars}
\begin{table}[htbp]
  \centering  
  \begin{tabular}{lcccccc}

  \toprule
  &
  $H_{\mathrm{node}}$&
  $H_{\mathrm{edge}}$&
  $H_{\mathrm{class}}$&
  $H_{\mathrm{agg}}^{\mathrm{M}}(\mathcal{G})$&
  $H_{\mathrm{edge}}^{*}$&
  LI
  \\

  \midrule

  









  cora&
  0.83& 
  0.81& 
  0.77& 
  0.99& 
  0.77& 
  0.59 

  \\

  citeseer&
  0.71& 
  0.74& 
  0.63& 
  0.97& 
  0.67& 
  0.45 

  \\

  pubmed&
  0.79& 
  0.80& 
  0.66& 
  0.94& 
  0.69& 
  0.41  

  \\
  roman-empire&
  0.05& 
  0.05& 
  0.02& 
  1.00& 
  -0.05& 
  0.11  
  \\

  amazon-ratings&
  0.38& 
  0.38& 
  0.13& 
  0.60& 
  0.14& 
  0.04  
  \\

  minesweeper&
  0.68& 
  0.68& 
  0.01& 
  0.61& 
  0.01& 
  0.00  
  \\

  tolokers&
  0.63& 
  0.59& 
  0.18& 
  0.00& 
  0.09& 
  0.01  
  \\

  questions&
  0.90&
  0.84&
  0.08&
  0.00&
  0.02&
  0.00
  \\

  squirrel-filtered&
  0.19& 
  0.21& 
  0.04& 
  0.00& 
  0.01& 
  0.00  
  \\

  chameleon-filtered&
  0.24& 
  0.24& 
  0.04& 
  0.25& 
  0.03& 
  0.01  
  \\
  
  \bottomrule
  \end{tabular}

  \caption{Heterophily levels of benchmark datasets. The $H_{\mathrm{agg}}^{\mathrm{M}}(\mathcal{G})$ stands for the aggregation homophily, calculated using $\hat{\mathbf{A}} = \mathbf{A} + \mathbf{I}$. The $H_{\mathrm{edge}}^{*}$ stands for the adjusted edge homophily, and the LI stands for the label informativeness. 
  The definitions of these metrics can be found in Appendix \ref{app:metric}.}
\end{table}
\label{table:heter_metric}

\subsection{Synthetic Datasets}
\label{appendix:syn-dataset}
In addition to the real-world datasets, we also tested parameterized diffusion on synthetic graphs generated with different homophily levels ranging from 0 to 1 using the method proposed in~\cite{abu2019mixhop}.
Here we give a review of the generation process.

More specifically, 
when generating the output graph $\mathcal{G}$ with a desired total number of nodes $N$, a total of $C$ classes,
and a homophily coefficient $\mu$,
the process begins by dividing the $N$ nodes into $C$ equal-sized classes.
Then the synthetic graph $\mathcal{G}$ (initially empty) is updated iteratively.
At each step,
a new node $v_i$ is added,
and its class $z_i$ is randomly assigned from the set $\{1, \ldots, C\}$.
Whenever a new node $v_i$ is added to the graph, we 
establish a connection between it and an existing node $v_j$ in $\mathcal{G}$ based on the probability $p_{ij}$ determined by the following rules:
\be
\label{eq:puv}
p_{ij} = \begin{cases}
d_j \times \mu, & \mbox{if $z_i = z_j$} \\
d_j \times (1-\mu) \times w_{d(z_i, z_j)}, & \mbox{otherwise}
\end{cases}.
\ee
where $z_i$ and $z_j$ are class labels of node $i$ and $j$ respectively, and $w_{d(z_i, z_j)}$ denotes the ``cost'' of connecting nodes from two distinct classes with a class distance of $d(z_i, z_j)$.
For a larger $\mu$, the chance of connecting with a node with the same label increases.
The distance between two classes simply implies the shortest distance between the two classes on a circle where classes are numbered from 1 to $C$. 
For instance, if $C = 6$, $z_i = 1$ and $z_j = 5$, then the distance between $z_i$ and $z_j$ is $2$. 
The weight exponentially decreases as the distance increases and is normalized such that $\sum_d w_d = 1$.
In addition, the probability $p_{ij}$ defined in Eq.~\eqref{eq:puv} is also normalized over the exiting nodes:
$$
    \bar{p}_{ij} = \frac{p_{ij}}{\sum_{k: v_k \in \mathcal{N}(v_i)} p_{ik}}
$$
Lastly, the features of each node in the output graph are sampled from overlapping 2D Gaussian distributions.
Each class has its own distribution defined separately.

\section{Hyperparameters}
\label{app:Hyperparameters}
The following table lists the optimal $\gamma$ and $\alpha$ used in models with the proposed methods.
\begin{table}[!ht]
    \centering
    \resizebox{1\hsize}{!}{
    \begin{tabular}{ccccccccc}
    \toprule
        ~ & ~ & roman-empire & amazon-ratings & minesweeper & tolokers & questions & squirrel-filtered & chameleon-iltered \\ \hline
        GCN (R(G)) & $\gamma$ & 0.1 & 0.3 & 0.1 & 0.9 & 0.1 & 0.1 & 0.1 \\ 
        ~ & $\alpha$ & 0.6 & 0.3 & 0.3 & 0.9 & 0.2 & 0 & 0 \\ \hline
        GAT (R(G)) &$\gamma$ & 0.8 & 0.2 & 0.2 & 0.5 & 0.5 & 0.2 & 0.1 \\ 
        ~ & $\alpha$ & 0.7 & 0 & 0.3 & 0.9 & 1 & 0.9 & 1 \\ \hline
        PD-GCN &$\gamma$ & 1 & 0.9 & 1 & 0.6 & 0.7 & 1 & 0.9 \\ 
        ~ & $\alpha$ & 0 & 0.9 & 1 & 0.4 & 0.8 & 0.1 & 0 \\ \hline
        PD-GAT &$\gamma$ & 0 & 0.9 & 0.4 & 1 & 0.1 & 0.4 & 0.6 \\ 
        ~ & $\alpha$ & 0 & 0.9 & 0.2 & 1 & 0.4 & 0.3 & 0.2 \\ \hline
        PD-GAT (R(G)) &$\gamma$ & 0.5 & 0.4 & 0.3 & 1 & 0.5 & 0.7 & 0.8 \\ 
        ~ & $\alpha$ & 0.9 & 1 & 0 & 0.7 & 1 & 0.4 & 0.5 \\ \hline
        PD-GAT-sep &$\gamma$ & 0.9 & 0.1 & 0.9 & 0.5 & 0.2 & 0.5 & 0.1 \\ 
        ~ & $\alpha$ & 0.4 & 0.9 & 0 & 1 & 0.2 & 0.3 & 0.9 \\ \hline
        PD-GAT-sep (R(G)) &$\gamma$ & 0.6 & 0.6 & 0.3 & 0.5 & 0.3 & 0.5 & 0.2 \\ 
        ~ & $\alpha$ & 0.8 & 0.2 & 0 & 0.9 & 0.4 & 0.9 & 0.9 \\ \bottomrule
    \end{tabular}
    }
    \caption{Optimal $\gamma$ and $\alpha$ for real-world datasets.}
\end{table}

For BernNet, LINKX, APPNP, $\omega$GCN and PGSO in real-world benchmark,
we perform a grid search for learning rate $\in \{0.01, 0.05, 0.1\}$, weight decay $\in \{0, 5e-7, 5e-6, 1e-5, 5e-5, 1e-4, 5e-4, 1e-3, 5e-3, 1e-2 \}$, dropout $\in \{0, 0.1, 0.3, 0.5, 0.7\}$. Model specific parameters are: (1) BernNet: the propagation steps $K = 10$; (2) LINKX: the numbder of layers of $\text{MLP}_{\text{A}}$ and $\text{MLP}_{\text{X}}$ in $\{1, 2\}$;
(3) APPNP: $\alpha \in \{ 0.1, 0.2, 0.5, 0.9 \}$ and up to $10^{th}$ power of the adjacency is used; (4) PGSO: initialization $\in$ \{ "GCN", "all zeros", "SymLaplacian", "RWLaplacian", "Adjacency"\}. 
We perform a grid search for the hyperparameters of DGN according to~\cite{beaini2021directional} for the learning rate in $\{ 10^{-5}, 10^{-4} \}$, the weight decay in $\{ 10^{-6}, 10^{-5}\}$, the dropout rate in $\{0.3, 0.5\}$, the aggregator in \{"mean-dir1-av", "mean-dir1-dx", "mean-dir1-av-dir1-dx" \}, the net type in \{"complex", "simple" \}.

\section{Comparison on Homophilic Datasets}
\label{app:homophily}
The following table summarizes the results on homophilic datasets, where we perform a grid search for learning rate $\in \{0.01, 0.05, 0.1\}$, weight decay $\in \{0, 5e-7, 5e-6, 1e-5, 5e-5, 1e-4, 5e-4, 1e-3, 5e-3, 1e-2 \}$, dropout $\in \{0, 0.1, 0.3, 0.5, 0.7\}$, and 64 hidden states. Note that the implementation of baselines follows~\cite{platonov2023critical}, where residual connections are adopted in each layer.
\begin{table}[!ht]
    \centering
    \begin{tabular}{cccc}
    \toprule
        ~ & Cora & Citeseer & Pubmed \\ \midrule
        ResNet & 72.03 $\pm$ 0.24 & 70.77 $\pm$ 1.81 & 88.01 $\pm$ 0.41 \\ 
        GCN & 86.15 $\pm$ 1.24 & 74.58 $\pm$ 0.97 & 89.63 $\pm$ 0.44 \\ 
        SAGE & 85.12 $\pm$ 1.64 & 74.47 $\pm$ 1.93 & 89.69 $\pm$ 0.51 \\ 
        GAT & 86.46 $\pm$ 1.02 & 74.13 $\pm$ 1.85 & 88.87 $\pm$ 0.65 \\ 
        GAT-sep & 84.24 $\pm$ 1.72 & 73.93 $\pm$ 1.93 & 89.14 $\pm$ 0.57 \\ 
        GT & 86.19 $\pm$ 0.99 & 74.23 $\pm$ 1.12 & 89.48 $\pm$ 0.52 \\ 
        GT-sep & 86.3 $\pm$ 1.13 & 74.14 $\pm$ 0.91 & 89.63 $\pm$ 0.52 \\ 
        \midrule
        DGN & 85.16 $\pm$ 1.17 & 72.70 $\pm$ 1.17 & 87.35 $\pm$ 0.53 \\ 
        $\omega$GCN & 86.13 $\pm$ 1.38 & 74.74 $\pm$ 1.39 & 88.65 $\pm$ 0.42 \\ 
        PGSO & 88.66 $\pm$ 0.94 & 76.55 $\pm$ 1.12 & OOM \\ 
        \midrule
        PD-GCN & 86.91 $\pm$ 1.45 & 75.17 $\pm$ 1.24 & 89.70 $\pm$ 0.45 \\ 
        PD-GAT & 85.40 $\pm$ 1.41 & 74.83 $\pm$ 1.75 & 89.48 $\pm$ 0.45 \\ \bottomrule
    \end{tabular}
    \caption{Experiments on homophily datasets proposed in~\cite{pei2020geom}.}
\end{table}

\section{Training}
\label{app:training}
In training and evaluating a model using a node classification benchmark dataset with $C$ distinct classes, 
each node $v_i \in \mathcal{V}$ has a label $z_i$ associated with it.
We denote $\mathbf{Z} \in \Rbb^{N \times C}$ as the one-hot encoding of labels.
Moreover, nodes are divided into three sets: the training set $\mathcal{V}_{\mathrm{train}}$, the validation set $\mathcal{V}_{\mathrm{val}}$ and the test set $\mathcal{V}_{\mathrm{test}}$.
In the training phase, 
the model uses features of all nodes under transductive learning.
The model only has access to labels of nodes in $\mathcal{V}_{\mathrm{train}}$ and in $\mathcal{V}_{\mathrm{val}}$ (for hyperparameter tuning),
while labels of nodes in $\mathcal{V}_{\mathrm{test}} = \mathcal{V} \setminus (\mathcal{V}_{\mathrm{train}} \cup \mathcal{V}_{\mathrm{val}})$ remain unknown to the model.

The cost function used in node classification tasks is the standard categorical cross-entropy loss~\cite{hamilton2020graph}, which is commonly used for multi-class classification tasks:
\be 
    \mathcal{L} = 
     - \frac{1}{|\mathcal{V}_{\mathrm{train}}|}\mathrm{trace}(\Z^\top \log \Y),
    \label{eq:loss}
\ee
where $\Y$ is the output from the model after $\mathrm{softmax}$
and $\log(\cdot)$ is applied element-wise.

\end{document}